\documentclass[11pt]{article}

\usepackage[preprint]{acl}

\usepackage{times}
\usepackage{latexsym}
\usepackage[T1]{fontenc}
\usepackage[utf8]{inputenc}
\usepackage{microtype}
\usepackage{inconsolata}
\usepackage{graphicx}
\usepackage{subcaption}
\usepackage{multirow}

\usepackage{amsthm}

\usepackage{amsmath,amssymb,mathtools}
\usepackage{booktabs}

\title{SPARE: State-Aware Recurrent Memory Compression for Efficient Hybrid Language Models}

\author{Ke Wan\thanks{Equal contribution.} \qquad Zaiyi Zheng\textsuperscript{*} \qquad Chen Chen \\
  \normalfont Department of Computer Science, University of Virginia \\
  \normalfont Charlottesville, VA, USA \\
  \normalfont\texttt{tbn5pj@virginia.edu}, \texttt{sjc4fq@virginia.edu}, \texttt{zrh6du@virginia.edu}}

\begin{document}
\maketitle

% Abstract is unnumbered under the official ACL style.
% \begin{abstract}
% Linear attention reduces the quadratic sequence-length complexity of softmax attention by maintaining fixed-size recurrent states.
% However, these states scale quadratically with the representation dimension and require repeated data transfers during decoding, creating a substantial GPU memory bottleneck.
% Recent studies have revealed significant low-rank redundancy in recurrent states, suggesting opportunities for more efficient representations.
% We introduce \texttt{SPARE} (\textbf{S}tate \textbf{P}rojection for \textbf{A}ccelerated \textbf{R}ecurrent \textbf{E}xecution), a training-free method that exploits this redundancy through fixed low-dimensional subspaces, heterogeneous rank allocation, and lightweight write compensation.
% By performing recurrent reads and updates directly on compact states, \texttt{SPARE} reduces memory traffic without online decomposition or dense-state reconstruction.
% \texttt{SPARE} achieves 2.53$\times$ full-model decoding speedup, 7.49$\times$ lower recurrent-kernel DRAM traffic, and 3.31$\times$ lower peak GPU memory, while preserving better associative-retrieval accuracy than structured pruning at matched memory budgets.
% \end{abstract}

\begin{abstract}
Modern hybrid language models increasingly incorporate recurrent architectures, including gated linear attention and selective state-space models, to improve inference efficiency.
While these architectures summarize historical information in fixed-size states, recurrent memory often exhibits substantial low-rank redundancy, offering opportunities for further compression.
To exploit this redundancy, we introduce \texttt{SPARE} (\textbf{S}tate \textbf{P}rojection for \textbf{A}ccelerated \textbf{R}ecurrent \textbf{E}xecution), a training-free framework that compresses recurrent memory into fixed low-dimensional subspaces.
\texttt{SPARE} constructs these subspaces through principal component analysis of recurrent-state statistics and supports rank allocation across layers and heads.
Crucially, \texttt{SPARE} maintains and updates recurrent states directly in the compact space, enabling efficient recurrent reads and writes without online decomposition or dense-state reconstruction.
This design reduces both recurrent-state memory traffic and per-request storage, improving decoding efficiency and supporting higher inference concurrency.
Across two hybrid language models, OLMo Hybrid 7B and Granite 4.0 H-Small, \texttt{SPARE} achieves $4\times$ recurrent-state compression and up to $2.00\times$ peak decoding throughput, while preserving or improving six-task average retrieval accuracy with its quality-oriented configuration.
\end{abstract}

% Main paper: standard numbered sections (1, 2, 3, ...).

\section{Introduction}
\label{sec:introduction}

Modern hybrid language models increasingly incorporate recurrent architectures, including gated linear attention \citep{yang2023gated,yang2025gated} and selective state-space models \citep{dao2024transformers,lieber2024jamba,ren2025samba}, to improve inference efficiency.
Unlike conventional softmax attention \citep{vaswani2017attention}, these recurrent modules summarize historical information into fixed-size states without maintaining a growing key-value cache.
However, eliminating the dependence on sequence length does not eliminate the cost of recurrent memory itself.
During decoding, repeatedly reading and updating large recurrent state matrices incurs substantial memory traffic, often making these operations memory-bandwidth-bound \citep{NEURIPS2022_67d57c32}.
Meanwhile, high-concurrency inference, a common strategy for improving serving throughput, requires maintaining separate recurrent states for concurrent requests.
Their accumulated memory footprint limits the batch size that can fit on a GPU, constraining achievable throughput.
Therefore, reducing recurrent-state memory is essential for both efficient decoding and high-throughput inference in hybrid language models.

\begin{figure*}[t]
    \centering
    \begin{subfigure}[c]{0.48\textwidth}
        \centering
        \includegraphics[width=\linewidth]{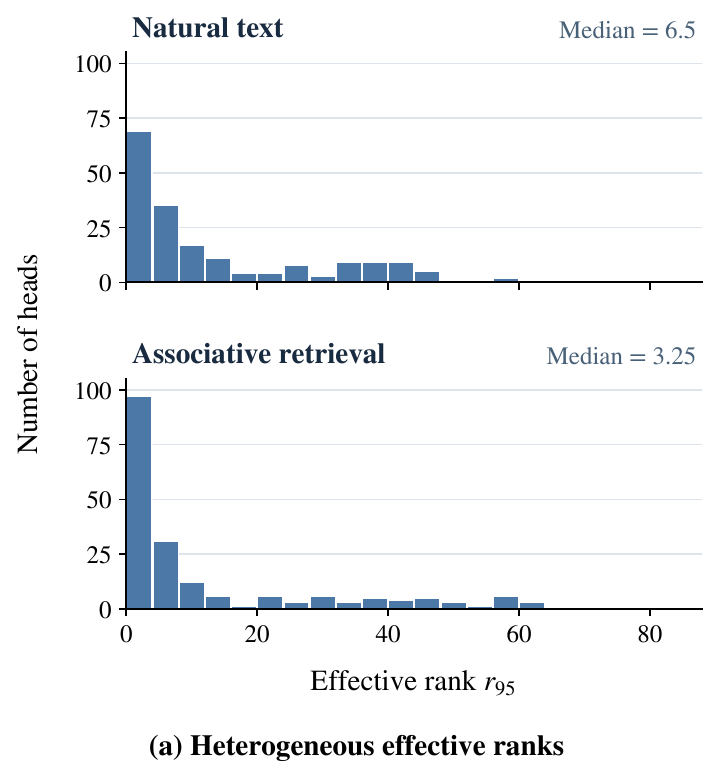}
        \phantomsubcaption
        \label{fig:spare_rank_distribution}
    \end{subfigure}
    \hfill
    \begin{subfigure}[c]{0.48\textwidth}
        \centering
        \includegraphics[width=\linewidth]{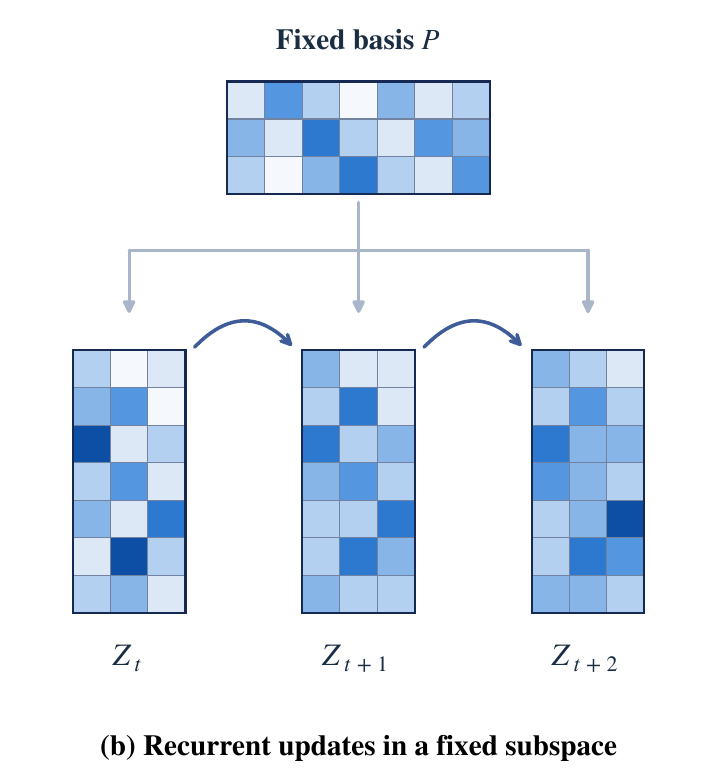}
        \phantomsubcaption
        \label{fig:spare_fixed_basis}
    \end{subfigure}
    \caption{
    \textbf{Motivation and intuition of SPARE.}
    \textbf{(a) Heterogeneous low-rank recurrent states.}
    The histograms show the distributions of effective ranks across 192 recurrent heads on natural text (top) and associative retrieval (bottom).
    For each head, we report the median $r_{95}$ across sampled states, where $r_{95}$ is the minimum rank retaining 95\% of the state's squared singular-value energy.
    Despite the original state dimension of $256 \times 256$, effective ranks are typically much smaller and vary substantially across heads.
    \textbf{(b) Recurrent updates in a fixed subspace.}
    SPARE represents the recurrent state as $\boldsymbol{S}_t \approx \boldsymbol{Z}_t\boldsymbol{P}^{\top}$, where the projection basis $\boldsymbol{P}$ remains fixed while the compact state $\boldsymbol{Z}_t$ evolves.
    Recurrent reads and updates operate directly on the compact state without reconstructing the dense matrix.
    }
    \label{fig:spare_motivation}
\end{figure*}

Reducing recurrent-state memory requires exploiting its underlying redundancy.
Prior studies have identified substantial low-rank structure in recurrent states \citep{nazari2026key}, with effective ranks varying considerably across layers and heads \citep{sun2026state}, as illustrated in Figure~\ref{fig:spare_rank_distribution}.
A natural approach is to reduce state dimensions through structured pruning of the original representation coordinates.
However, low-rank structure does not imply that important information is concentrated in a small subset of these coordinates.
Consequently, aggressive pruning may discard information essential to associative memory.

Low-rank projection offers a natural alternative by preserving the dominant directions of recurrent states rather than restricting compression to the original coordinates.
With an appropriately chosen subspace, it can minimize state reconstruction error under a fixed rank budget.
Importantly, recurrent memory is shaped by successive state updates, so its dominant directions need not coincide with those of the inputs that construct it.
This motivates deriving compression subspaces directly from recurrent-state statistics rather than input representations alone.
Moreover, performing recurrent reads and updates directly in the resulting compact representation can substantially reduce both memory traffic and per-request state storage.
However, the heterogeneous memory requirements across layers and heads make effective subspace selection and capacity allocation essential.

To address these challenges, we introduce \texttt{SPARE} (\textbf{S}tate \textbf{P}rojection for \textbf{A}ccelerated \textbf{R}ecurrent \textbf{E}xecution), a training-free framework for efficient recurrent memory compression.
\texttt{SPARE} constructs fixed low-dimensional subspaces from recurrent-state statistics using offline principal component analysis.
It further exploits heterogeneous memory requirements through rank allocation across layers and heads, while also supporting uniform allocation for efficient execution.
As illustrated in Figure~\ref{fig:spare_fixed_basis}, \texttt{SPARE} maintains recurrent states directly in compact subspaces, enabling native recurrent reads and updates without reconstructing the original dense states.

Our contributions are summarized as follows:
\begin{itemize}
    \item We investigate the heterogeneous low-rank structure of recurrent memory and identify state-aware subspace construction and flexible rank allocation as complementary opportunities for efficient compression.

    \item We propose \texttt{SPARE}, a training-free framework that combines state-derived projection, flexible rank allocation, and compact recurrent execution across different recurrent architectures.

    \item We achieve $4\times$ recurrent-state compression and up to $2.00\times$ peak decoding throughput on hybrid language models, while preserving or improving six-task average cloze-retrieval accuracy with our quality-oriented configuration.
\end{itemize}

\section{Background and Motivation}
\label{sec:background}

\subsection{Recurrent Memory in Hybrid Language Models}
\label{sec:recurrent_memory}

Modern hybrid language models incorporate recurrent architectures, including gated linear attention \citep{yang2025gated} and selective state-space models \citep{dao2024transformers}, to efficiently represent historical information.
Despite their different update mechanisms, both architectures maintain matrix-valued recurrent states that are updated and accessed during sequence processing.
For each recurrent layer $\ell$ and head $h$, we denote the state at position $t$ by $\boldsymbol{S}_t^{(\ell,h)} \in \mathbb{R}^{d_v \times d_s}$, where $d_s$ is the recurrent-state dimension and $d_v$ is the corresponding feature dimension.
Their recurrent computation can be expressed as
\begin{equation}
\begin{aligned}
\boldsymbol{S}_t &= \mathcal{F}_t(\boldsymbol{S}_{t-1},\boldsymbol{x}_t),\\
\boldsymbol{o}_t &= \mathcal{R}_t(\boldsymbol{S}_t,\boldsymbol{x}_t),
\end{aligned}
\label{eq:recurrent_formulation}
\end{equation}
where $\mathcal{F}_t$ and $\mathcal{R}_t$ denote the architecture-specific state update and readout operations, respectively.
For clarity, we omit layer and head indices when discussing individual recurrent states.

Unlike conventional attention, recurrent-state storage is independent of sequence length, but each decoding step requires accessing and updating the state.
For a batch of $B$ concurrent requests, the total recurrent-state storage scales as
\begin{equation}
M_{\mathrm{state}}
=
O\left(
B\sum_{\ell,h}d_v^{(\ell,h)}d_s^{(\ell,h)}
\right).
\label{eq:recurrent_memory_complexity}
\end{equation}
Consequently, large recurrent states increase memory traffic during decoding and limit the batch capacity available for high-throughput inference.

\subsection{Recurrent-State Redundancy}
\label{sec:recurrent_state_redundancy}

Although a recurrent state $\boldsymbol{S}_t$ contains $d_vd_s$ elements, its effective dimensionality can be substantially smaller.
Let $\sigma_{t,1}\geq\cdots\geq\sigma_{t,m}$ denote its singular values, where $m=\min(d_v,d_s)$.
We define the effective rank required to retain a fraction $\eta\in(0,1)$ of the state's spectral energy as
\begin{equation}
r_{\eta}(\boldsymbol{S}_t)
=
\min\left\{
r:
\frac{\sum_{i=1}^{r}\sigma_{t,i}^{2}}
{\sum_{i=1}^{m}\sigma_{t,i}^{2}}
\geq\eta
\right\}.
\label{eq:effective_rank}
\end{equation}
Prior studies have observed substantial low-rank structure in recurrent states \citep{nazari2026key}, with effective ranks varying considerably across layers and heads \citep{sun2026state}.
As illustrated in Figure~\ref{fig:spare_rank_distribution}, this redundancy suggests that recurrent memory can be represented using substantially fewer dimensions.

For any state matrix $\boldsymbol{S}_t$, the Eckart--Young theorem establishes that its optimal rank-$r$ approximation minimizes the Frobenius reconstruction error \citep{eckart1936approximation}:
\begin{equation}
\min_{\operatorname{rank}(\widehat{\boldsymbol{S}})\leq r}
\|\boldsymbol{S}_t-\widehat{\boldsymbol{S}}\|_F^2
=
\sum_{i=r+1}^{m}\sigma_{t,i}^{2}.
\label{eq:optimal_rank_approximation}
\end{equation}
Unlike coordinate pruning, low-rank approximation can preserve information distributed across arbitrary representation directions.
However, independently decomposing recurrent states at every step would introduce substantial overhead, while reconstructing dense states would diminish the benefits of compression.
Exploiting this redundancy therefore requires a fixed compact representation that supports efficient recurrent computation while accommodating heterogeneous memory requirements.

\section{SPARE: Compact Recurrent States}
\label{sec:method}

SPARE exploits recurrent-state redundancy by maintaining memory in fixed low-dimensional subspaces and performing recurrent computation directly on compact states.
The projection bases and rank allocations are determined through offline calibration and remain fixed during inference.

\subsection{State-Aware Low-Rank Representation}
\label{sec:state_aware_representation}

Motivated by the low-rank structure of recurrent memory, SPARE compresses the recurrent-state dimension while preserving the full feature dimension.
For clarity, we consider a fixed orthonormal basis $\boldsymbol{P}\in\mathbb{R}^{d_s\times r}$, where $r\ll d_s$ and $\boldsymbol{P}^{\top}\boldsymbol{P}=\boldsymbol{I}_r$.
Instead of maintaining the original recurrent state $\boldsymbol{S}_t$, SPARE represents it using a compact coefficient matrix $\boldsymbol{Z}_t\in\mathbb{R}^{d_v\times r}$:
\begin{equation}
\boldsymbol{S}_t
\approx
\widehat{\boldsymbol{S}}_t
=
\boldsymbol{Z}_t\boldsymbol{P}^{\top}.
\label{eq:spare_representation}
\end{equation}
This representation reduces recurrent-state storage from $O(d_vd_s)$ to $O(d_vr)$ without requiring dense-state reconstruction.

SPARE constructs the fixed projection basis directly from recurrent-state statistics.
For each layer $\ell$ and head $h$, we collect recurrent states from calibration sequences and compute the uncentered second-moment matrix
\begin{equation}
\boldsymbol{\Sigma}_{\ell,h}
=
\frac{1}{N}
\sum_{i=1}^{N}
\left(\boldsymbol{S}_i^{(\ell,h)}\right)^{\top}
\boldsymbol{S}_i^{(\ell,h)}.
\label{eq:state_covariance}
\end{equation}
The projection basis $\boldsymbol{P}_{\ell,h}$ consists of the leading $r_{\ell,h}$ eigenvectors of $\boldsymbol{\Sigma}_{\ell,h}$.
Equivalently, it minimizes the average reconstruction error among fixed orthogonal projections of rank $r_{\ell,h}$:
\begin{equation}
\boldsymbol{P}_{\ell,h}^{*}
=
\arg\min_{\boldsymbol{P}^{\top}\boldsymbol{P}=\boldsymbol{I}}
\mathbb{E}_{\mathrm{cal}}
\left[
\left\|
\boldsymbol{S}_t^{(\ell,h)}
-
\boldsymbol{S}_t^{(\ell,h)}
\boldsymbol{P}\boldsymbol{P}^{\top}
\right\|_F^2
\right].
\label{eq:state_pca_objective}
\end{equation}
Unlike input-derived PCA, this procedure directly identifies the dominant directions of accumulated recurrent memory.
The resulting bases remain fixed throughout inference, avoiding online matrix decomposition.

\subsection{Rank Allocation}
\label{sec:rank_allocation}

Since recurrent states exhibit different degrees of redundancy across layers and heads, SPARE supports heterogeneous and uniform rank allocation under a fixed memory budget.

\paragraph{Heterogeneous Allocation (SPARE-H).}
For each layer $\ell$ and head $h$, we estimate the normalized spectral contribution of the $j$-th singular direction from calibration states:
\begin{equation}
w_{\ell,h,j}
=
\mathbb{E}_{\mathrm{cal}}
\left[
\frac{
\sigma_j^2\left(\boldsymbol{S}_t^{(\ell,h)}\right)
}{
\left\|\boldsymbol{S}_t^{(\ell,h)}\right\|_F^2
}
\right],
\label{eq:spectral_contribution}
\end{equation}
where $\sigma_j(\boldsymbol{S}_t^{(\ell,h)})$ denotes the $j$-th largest singular value of the recurrent state.
Given a total rank budget $R$, we greedily allocate additional dimensions to heads with the largest remaining spectral contributions, subject to predefined rank bounds and $\sum_{\ell,h}r_{\ell,h}=R$.
The resulting allocation assigns different ranks to individual heads while maintaining the prescribed total state capacity.

\paragraph{Uniform Allocation (SPARE-U).}
Alternatively, SPARE assigns the same rank to every recurrent head under the same total memory budget.
Both configurations use the same state-aware projection procedure, with uniform allocation providing regular state dimensions for efficient GPU execution.

\subsection{Compact Recurrent Updates}
\label{sec:compact_recurrent_updates}

With the fixed projection basis $\boldsymbol{P}$, SPARE performs recurrent reads and updates directly on the compact coefficient matrix $\boldsymbol{Z}_t$.
The corresponding read and write vectors are projected into the retained subspace while preserving the structure of the native recurrent operations.

\paragraph{Gated Linear Attention.}
For Gated DeltaNet \citep{yang2025gated}, the recurrent state is updated through the Gated Delta Rule:
\begin{equation}
\begin{aligned}
\boldsymbol{S}_t
&=
\alpha_t\boldsymbol{S}_{t-1}
+
\beta_t
\left(
\boldsymbol{v}_t
-
\alpha_t\boldsymbol{S}_{t-1}\boldsymbol{k}_t
\right)
\boldsymbol{k}_t^{\top},\\
\boldsymbol{o}_t
&=
\boldsymbol{S}_t\boldsymbol{q}_t,
\end{aligned}
\label{eq:gated_delta_rule}
\end{equation}
where $\alpha_t$ controls state decay and $\beta_t$ determines the strength of memory updates.
SPARE projects the key and query vectors into the fixed subspace:
\begin{equation}
\widehat{\boldsymbol{k}}_t
=
\boldsymbol{P}^{\top}\boldsymbol{k}_t,
\qquad
\widehat{\boldsymbol{q}}_t
=
\boldsymbol{P}^{\top}\boldsymbol{q}_t.
\label{eq:spare_gdn_projection}
\end{equation}
The compact recurrent update and readout are then given by
\begin{equation}
\begin{aligned}
\boldsymbol{Z}_t
&=
\alpha_t\boldsymbol{Z}_{t-1}
+
\beta_t
\left(
\boldsymbol{v}_t
-
\alpha_t\boldsymbol{Z}_{t-1}
\widehat{\boldsymbol{k}}_t
\right)
\widehat{\boldsymbol{k}}_t^{\top},\\
\widehat{\boldsymbol{o}}_t
&=
\boldsymbol{Z}_t\widehat{\boldsymbol{q}}_t.
\end{aligned}
\label{eq:spare_gdn_update}
\end{equation}
Both the recurrent update and readout operate directly on the compact state without reconstructing the original key dimension.

\paragraph{Selective State-Space Models.}
For Mamba-2 \citep{dao2024transformers}, the matrix-valued recurrent state can be expressed in a compatible orientation as
\begin{equation}
\begin{aligned}
\boldsymbol{S}_t
&=
a_t\boldsymbol{S}_{t-1}
+
\boldsymbol{u}_t\boldsymbol{B}_t^{\top},\\
\boldsymbol{o}_t
&=
\boldsymbol{S}_t\boldsymbol{C}_t,
\end{aligned}
\label{eq:mamba2_recurrence}
\end{equation}
where $a_t$ denotes the state decay, $\boldsymbol{u}_t$ is the input-dependent feature vector, and $\boldsymbol{B}_t,\boldsymbol{C}_t\in\mathbb{R}^{d_s}$ specify the state-input and readout directions.
Using the fixed projection basis, SPARE computes
\begin{equation}
\widehat{\boldsymbol{B}}_t
=
\boldsymbol{P}^{\top}\boldsymbol{B}_t,
\qquad
\widehat{\boldsymbol{C}}_t
=
\boldsymbol{P}^{\top}\boldsymbol{C}_t.
\label{eq:spare_mamba_projection}
\end{equation}
The compact recurrence becomes
\begin{equation}
\begin{aligned}
\boldsymbol{Z}_t
&=
a_t\boldsymbol{Z}_{t-1}
+
\boldsymbol{u}_t\widehat{\boldsymbol{B}}_t^{\top},\\
\widehat{\boldsymbol{o}}_t
&=
\boldsymbol{Z}_t\widehat{\boldsymbol{C}}_t.
\end{aligned}
\label{eq:spare_mamba_update}
\end{equation}
This formulation preserves the native state-transition structure while reducing the recurrent-state dimension from $d_s$ to $r$.
For both architectures, the compact recurrence introduces no state-projection error when the initial state and all subsequent state-writing directions lie within the retained subspace under identical inputs and transition parameters.

\subsection{Efficient Decoding}
\label{sec:efficient_decoding}

SPARE maintains compact recurrent states throughout decoding, reducing the per-head state size from $d_vd_s$ to $d_vr$.
For SPARE-H, the variable-rank coefficient matrices are packed into contiguous GPU buffers, with each head accessing its compact state through fixed offsets.
For SPARE-U, the uniform ranks enable regular state layouts and optimized fixed-width recurrent kernels.
Both implementations perform recurrent reads and updates directly on compact states without dense-state reconstruction.

For a head of rank $r$, recurrent-state read/write traffic scales as $O(d_vr)$ rather than $O(d_vd_s)$, excluding projection and auxiliary overheads.
The fixed projection bases are shared across requests, while each request maintains its own compact recurrent states.
Under a fixed GPU memory budget, this reduction also enables larger concurrent batches and higher decoding throughput.
We implement the recurrent operations using GPU kernels and maintain address-stable states to support CUDA Graph execution.
\section{Experiments}
\label{sec:experiments}
% Suggested topics: setup, quality-efficiency tradeoffs, DRAM traffic,
% decode latency, full-request performance, concurrency, and ablations.

\begin{table*}[t]
    \centering
    \caption{
    \textbf{Quality and efficiency of recurrent-state compression on hybrid language models.}
    We compare Dense, DRRQR, SPARE-H, and SPARE-U under matched $4\times$ recurrent-state compression budgets.
    Task results report contains Accuracy (\%), with \textbf{Avg.} denoting the unweighted mean of six downstream tasks, excluding synthetic associative retrieval.
    State denotes recurrent-state storage per request, while Peak Decode reports full-model peak decoding throughput relative to Dense.
    DRRQR is evaluated only on OLMo Hybrid 7B.
    Bold values indicate the best result within each backbone.
    }
    \label{tab:main_results}
    \small
    \setlength{\tabcolsep}{5pt}
    \renewcommand{\arraystretch}{1.12}
    \resizebox{\textwidth}{!}{
    \begin{tabular}{@{}l c c c c c c c@{}}
        \toprule
        &
        \multicolumn{4}{c}{\textbf{OLMo Hybrid 7B (GDN)}} &
        \multicolumn{3}{c}{\textbf{Granite 4.0 H-Small (Mamba-2)}} \\
        \cmidrule(lr){2-5}
        \cmidrule(lr){6-8}

        \textbf{Benchmark} &
        \textbf{Dense} &
        \textbf{DRRQR} &
        \textbf{SPARE-H} &
        \textbf{SPARE-U} &
        \textbf{Dense} &
        \textbf{SPARE-H} &
        \textbf{SPARE-U} \\
        \midrule

        Associative Retrieval
        & 97.66 & 97.66 & \textbf{98.24} & 96.48
        & \textbf{99.61} & \textbf{99.61} & 97.46 \\

        \addlinespace[3pt]

        SWDE
        & 74.23 & 46.11 & \textbf{75.35} & 74.41
        & 76.48 & \textbf{79.94} & 73.48 \\

        SQuAD
        & 58.29 & 42.24 & \textbf{64.73} & 60.44
        & 55.16 & \textbf{63.90} & 50.23 \\

        FDA
        & 78.66 & 57.31 & \textbf{78.93} & 78.66
        & 82.20 & \textbf{83.20} & 78.47 \\

        TriviaQA
        & \textbf{73.70} & 65.52 & 72.04 & 72.16
        & \textbf{74.82} & 73.28 & 72.69 \\

        Natural Questions
        & \textbf{40.89} & 34.68 & 40.23 & 39.78
        & 45.39 & \textbf{46.12} & 42.32 \\

        DROP
        & 34.83 & 28.37 & \textbf{38.00} & 34.64
        & 33.16 & \textbf{35.12} & 30.28 \\

        \cmidrule(lr){1-8}

        \textbf{Avg. (\%) $\uparrow$}
        & 60.10 & 45.71 & \textbf{61.55} & 60.02
        & 61.20 & \textbf{63.59} & 57.91 \\

        \midrule

        \textbf{State (MiB) $\downarrow$}
        & 50.625 & \textbf{12.656} & \textbf{12.656} & \textbf{12.656}
        & 144 & \textbf{36} & \textbf{36} \\

        \textbf{Peak Decode ($\times$) $\uparrow$}
        & 1.000 & \textbf{1.561} & 1.431 & 1.532
        & 1.000 & 1.547 & \textbf{1.995} \\

        \bottomrule
    \end{tabular}
    }
\end{table*}

\subsection{Experimental Setup}
\label{sec:experimental_setup}

\paragraph{Models.}
We evaluate \texttt{SPARE} on two pretrained hybrid language models: OLMo Hybrid 7B \citep{merrill2026olmo} (Gated DeltaNet and softmax attention) and Granite 4.0 H-Small (32B total and 9B activated parameters, combining Mamba-2, softmax attention, and MoE).
Model weights use BF16, while recurrent states use FP32.
Both \texttt{SPARE-H} and \texttt{SPARE-U} are training-free and employ offline State PCA calibration in our main experiments.

\paragraph{Baselines.}
We compare the uncompressed models (\textbf{Dense}) with \textbf{SPARE-H} (heterogeneous ranks) and \textbf{SPARE-U} (uniform ranks), both using $4\times$ recurrent-state compression.
We additionally evaluate structured channel pruning (\textbf{DRRQR}) \citep{nazari2026key} on OLMo Hybrid and input-derived PCA in our ablation studies.
No recovery fine-tuning is applied.

\paragraph{Evaluation Tasks.}
We evaluate six retrieval and question-answering benchmarks: SWDE \citep{hao2011one}, SQuAD \citep{rajpurkar-etal-2016-squad}, FDA \citep{arora2024simple}, TriviaQA \citep{joshi-etal-2017-triviaqa}, Natural Questions \citep{kwiatkowski-etal-2019-natural}, and DROP \citep{dua-etal-2019-drop}, using Contains Accuracy with up to 48 greedy decoding tokens.
We additionally evaluate 512 synthetic associative-retrieval examples.
For long-context evaluation, we use six RULER tasks \citep{hsieh2024ruler} at 4K, 16K, 32K, and 64K on OLMo Hybrid, with 128 paired examples per task and length.

\begin{table}[t]
    \centering
    \caption{
    \textbf{Batch capacity and decoding efficiency on A100 GPUs.}
    All SPARE configurations use $4\times$ recurrent-state compression.
    Batch Gain measures the increase in maximum feasible batch size relative to Dense.
    Decode Speedup is measured at the Dense maximum batch size ($B=256$ for OLMo and $B=352$ for Granite).
    }
    \label{tab:batch_scaling}
    \small
    \setlength{\tabcolsep}{4pt}
    \renewcommand{\arraystretch}{1.12}
    \resizebox{0.94\columnwidth}{!}{
    \begin{tabular}{@{}llcc@{}}
        \toprule
        \textbf{Model} &
        \textbf{Method} &
        \textbf{Batch Gain} &
        \textbf{Decode Speedup} \\
        \midrule

        \multirow{2}{*}{\textbf{OLMo Hybrid}}
        & \textbf{SPARE-H} & 2.00$\times$ & 1.280$\times$ \\
        & \textbf{SPARE-U} & \textbf{2.13$\times$} & \textbf{1.385$\times$} \\

        \addlinespace[3pt]

        \multirow{2}{*}{\textbf{Granite H-Small}}
        & \textbf{SPARE-H} & 2.55$\times$ & 1.219$\times$ \\
        & \textbf{SPARE-U} & \textbf{3.64$\times$} & \textbf{1.426$\times$} \\

        \bottomrule
    \end{tabular}
    }
\end{table}

\paragraph{System Evaluation.}
We benchmark full-model decoding on NVIDIA A100 40GB GPUs (one for OLMo Hybrid and four for Granite), using a 128-token prefix, 64 decoding steps, and CUDA Graphs.
We report matched-batch and peak decoding throughput, recurrent-state memory per request, and batch-capacity gains relative to Dense.

\begin{figure}[t]
    \centering
    \includegraphics[width=\columnwidth]{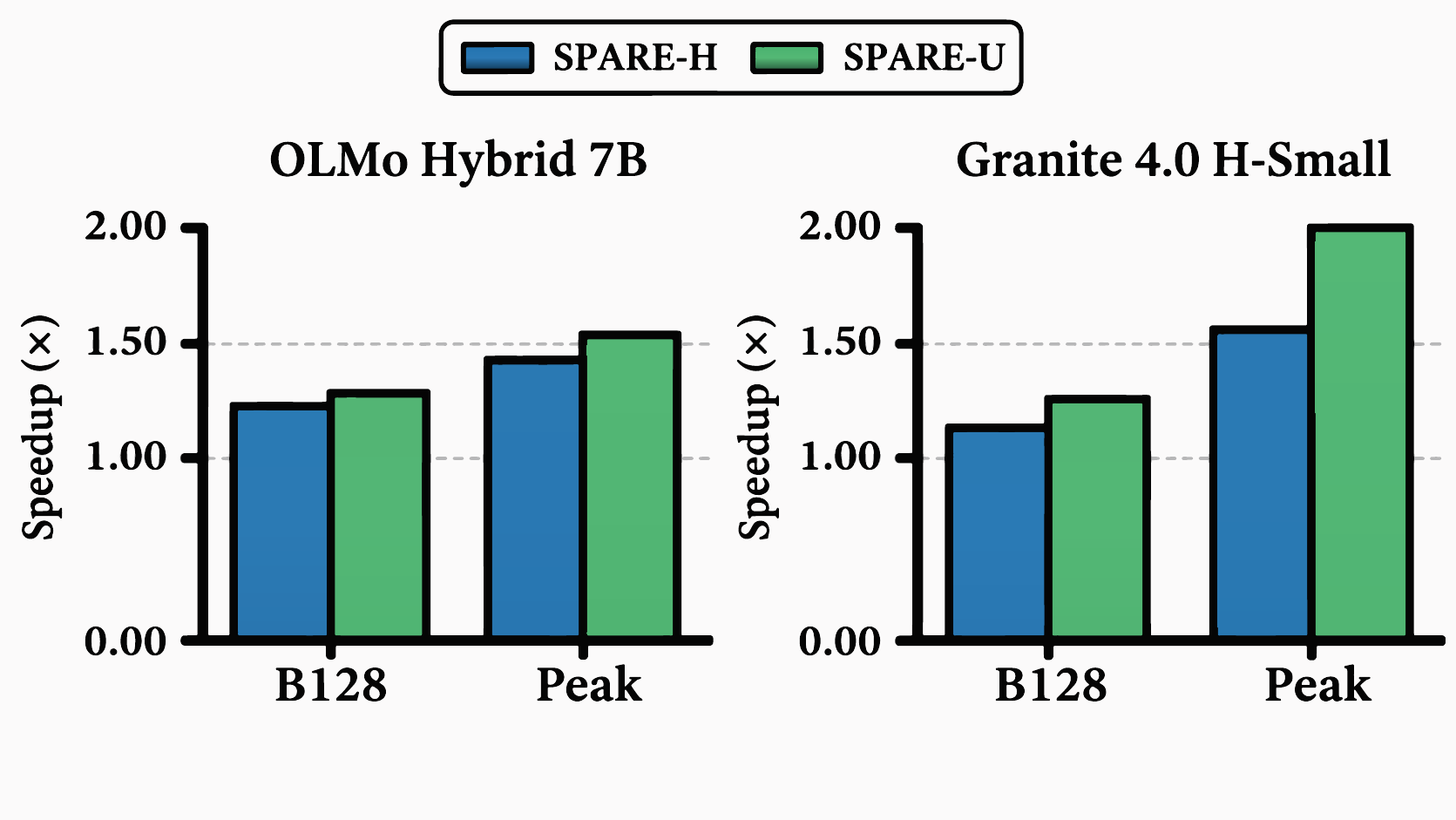}
    
    \caption{
    \textbf{Full-model decoding efficiency on hybrid language models.}
    We compare \texttt{SPARE-H} and \texttt{SPARE-U} under $4\times$ recurrent-state compression, using 128-token prefixes and 64 decoding steps.
    \textbf{B128} measures throughput at matched batch size, while \textbf{Peak} reports the highest throughput across feasible batches, capturing both execution efficiency and concurrency benefits.
    }

    \label{fig:decode_efficiency}
\end{figure}

\subsection{Task-dependent Performance}
\label{sec:task_performance}

Table~\ref{tab:main_results} shows that substantial recurrent-state compression does not necessarily compromise downstream retrieval performance.
Despite a fourfold reduction in state storage, \texttt{SPARE-H} preserves or slightly improves average retrieval accuracy on both hybrid language models, suggesting considerable redundancy in their recurrent representations.
However, performance varies across tasks, indicating that different retrieval capabilities exhibit different sensitivities to compression.
At matched memory budgets, \texttt{SPARE-H} consistently outperforms \texttt{SPARE-U} in average accuracy, demonstrating the benefits of allocating limited state capacity according to heterogeneous memory requirements.
Moreover, the substantial performance gap between \texttt{SPARE} and DRRQR highlights the advantage of preserving informative low-dimensional subspaces rather than pruning original representation coordinates.
Together, these results suggest that effective recurrent-state compression depends not only on the retained state size, but also on how the compressed representation and its capacity are selected.

\begin{table}[t]
    \centering
    \caption{
    \textbf{Comparison of projection subspaces.}
    Six-task average Contains Accuracy (\%) under matched compression budgets and heterogeneous rank allocations.
    }
    \label{tab:subspace_ablation}
    \small
    \setlength{\tabcolsep}{6pt}
    \renewcommand{\arraystretch}{1.12}
    \begin{tabular}{@{}lcc@{}}
        \toprule
        \textbf{Subspace} &
        \textbf{OLMo Hybrid} &
        \textbf{Granite H-Small} \\
        \midrule
        \textbf{Input PCA} & 61.32 & 62.22 \\
        \textbf{State PCA} & \textbf{61.55} & \textbf{63.59} \\
        \bottomrule
    \end{tabular}
\end{table}

\begin{figure}[t]
    \centering
    \includegraphics[width=\columnwidth]{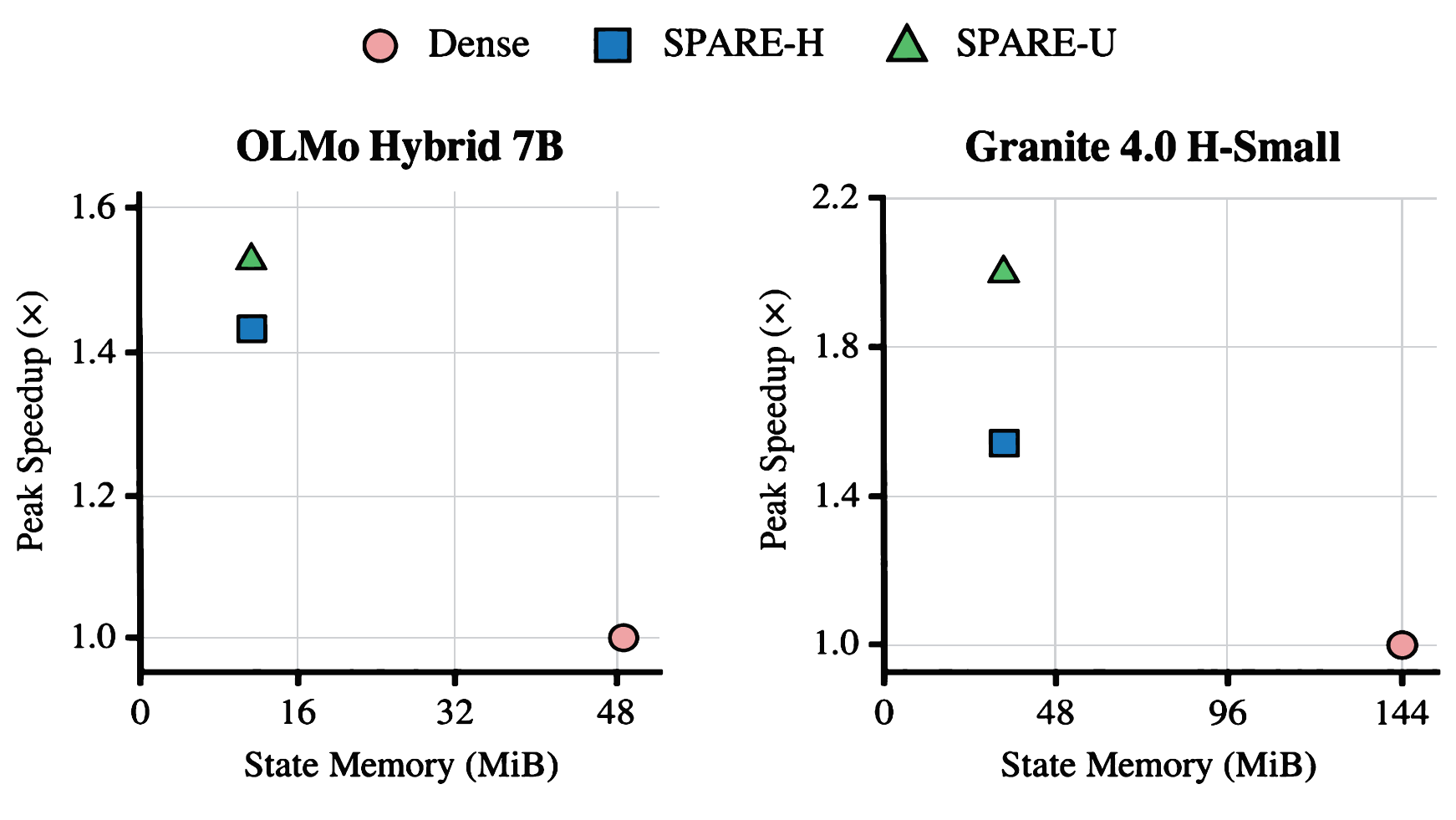}
    \caption{
    \textbf{Memory--throughput trade-off under recurrent-state compression.}
    We compare per-request recurrent-state storage and peak full-model decoding speedup relative to Dense on two hybrid language models.
    Both \texttt{SPARE-H} and \texttt{SPARE-U} reduce state storage by $4\times$, while their different peak throughput gains reflect the impact of rank allocation and GPU execution efficiency.
    }
    \label{fig:memory_throughput}
\end{figure}

\begin{figure*}[t]
    \centering
    \begin{subfigure}[t]{0.49\textwidth}
        \centering
        \includegraphics[width=\linewidth]{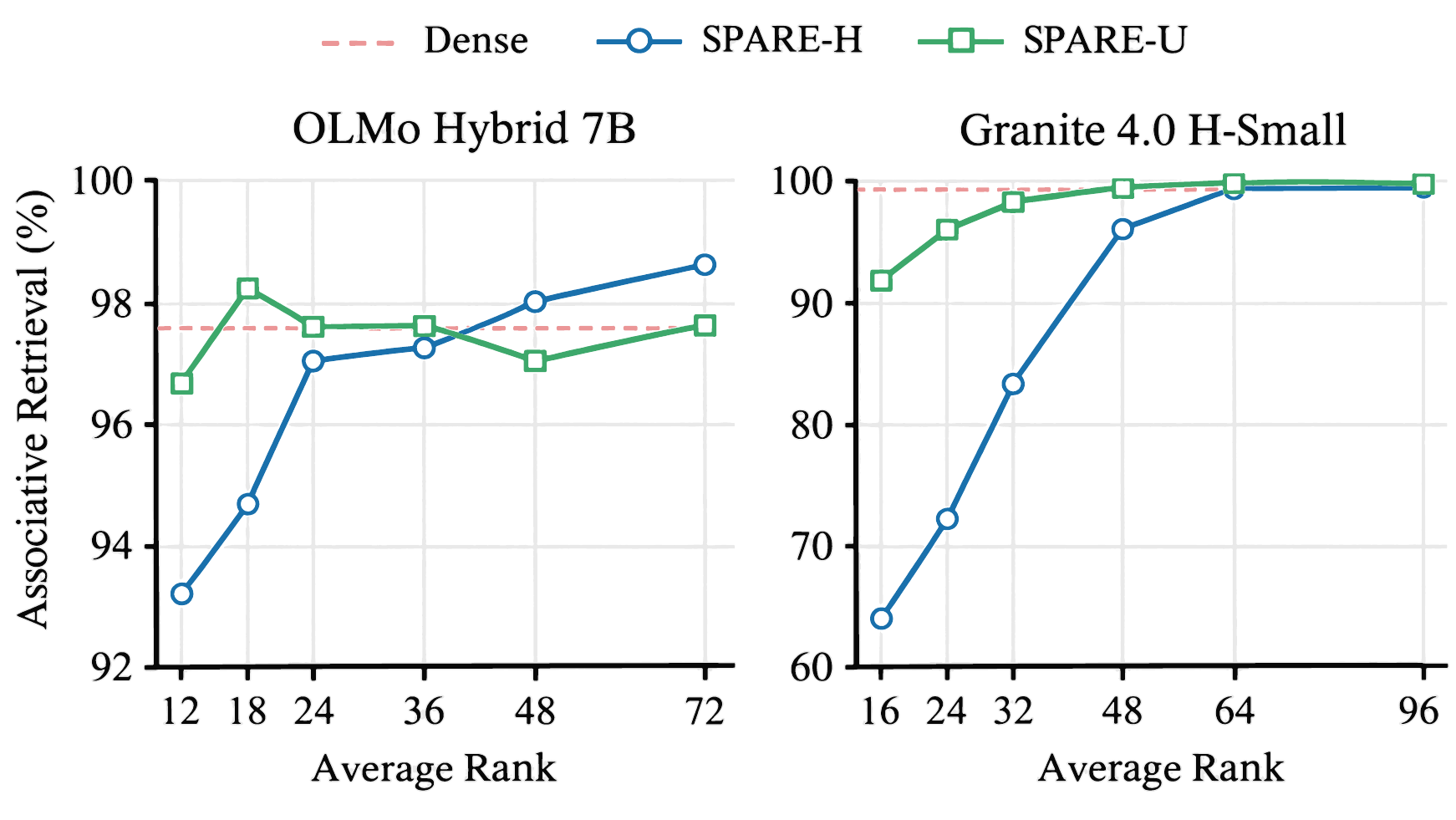}
        \caption{Associative retrieval accuracy.}
        \label{fig:rank_associative}
    \end{subfigure}
    \hfill
    \begin{subfigure}[t]{0.49\textwidth}
        \centering
        \includegraphics[width=\linewidth]{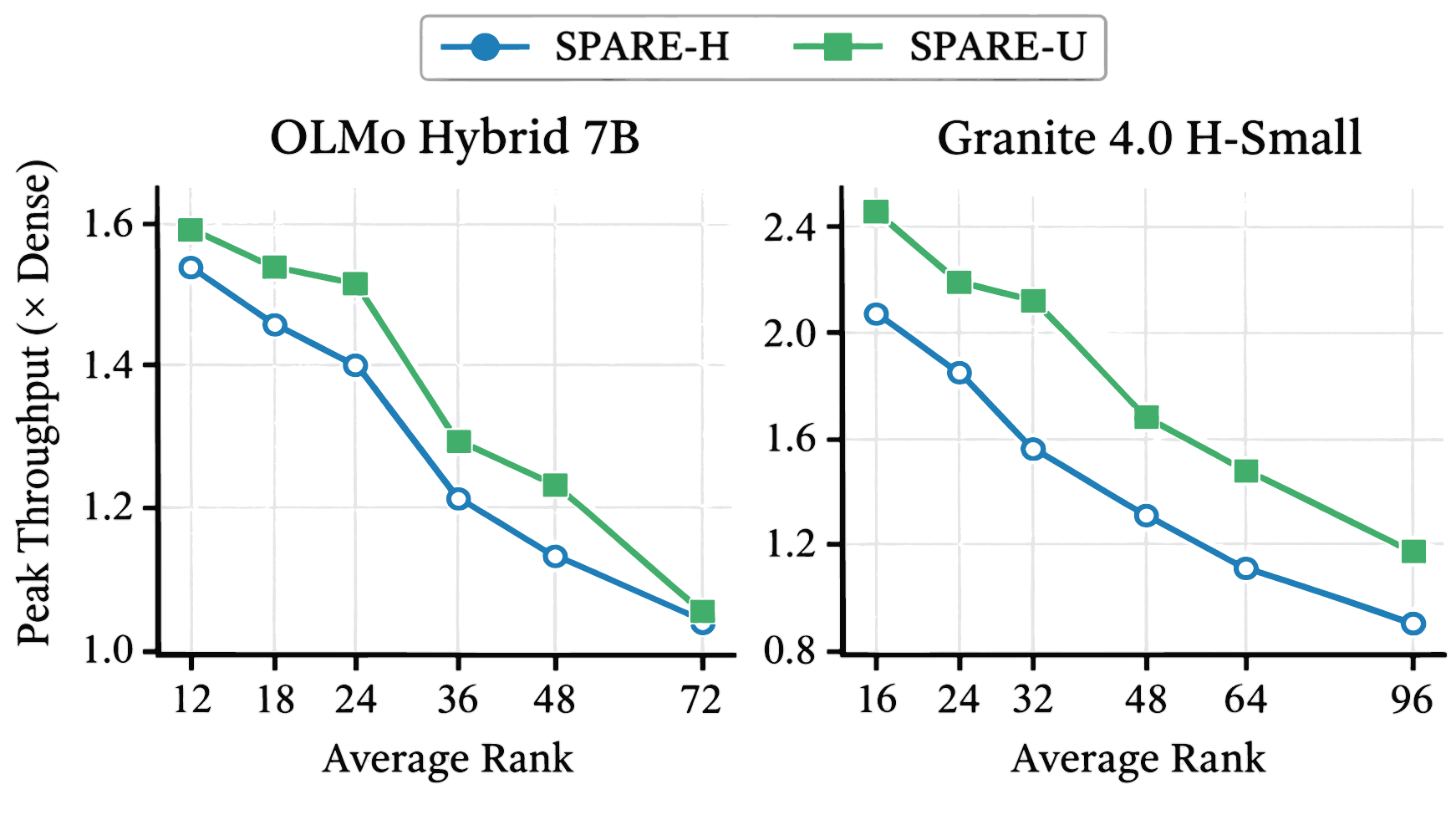}
        \caption{Peak decoding throughput.}
        \label{fig:rank_throughput}
    \end{subfigure}
    \caption{
    \textbf{Rank-dependent quality--efficiency trade-offs on hybrid language models.}
    Increasing the retained rank generally improves associative retrieval while reducing peak decoding throughput.
    Heterogeneous ranks denote the average rank across heads, while uniform ranks are shared across heads.
    These sweeps use the original input-derived projection configurations.
    }
    \label{fig:rank_ablation}
\end{figure*}

\subsection{Subspace Ablation}
\label{sec:subspace_ablation}

Table~\ref{tab:subspace_ablation} compares State PCA with input-derived PCA under identical rank allocations and compression budgets.
State PCA achieves higher average retrieval accuracy on both hybrid backbones, with a more pronounced improvement on Granite, while the difference on OLMo is not statistically significant.
These results suggest that constructing projection bases from accumulated recurrent states can better preserve task-relevant information than relying solely on input statistics.
However, the varying improvements across architectures highlight that minimizing state reconstruction error does not necessarily optimize downstream performance.

\subsection{Efficiency Analysis}
\label{sec:inference_efficiency}

Figure~\ref{fig:decode_efficiency} shows that both \texttt{SPARE} variants accelerate full-model decoding on A100 GPUs.
The primary benefit comes from reducing recurrent-update costs, partially offset by the overhead of projecting inputs into compact subspaces.
On Granite H-Small at batch size 128, \texttt{SPARE-U} saves 21.51 ms in recurrent updates with only 1.28 ms of projection overhead, reducing overall decoding step time by 20.71 ms.
At matched $4\times$ compression budgets, \texttt{SPARE-U} achieves higher throughput than \texttt{SPARE-H}, as uniform ranks enable fixed-width recurrent kernels with more regular memory access and higher effective bandwidth, while avoiding the additional projection overhead of heterogeneous execution.
Figure~\ref{fig:memory_throughput} and Table~\ref{tab:batch_scaling} further show that compact states increase batch capacity, reaching $3.64\times$ capacity and $1.426\times$ matched-batch speedup on Granite.
However, larger batches increase per-step latency, so peak throughput gains are smaller than the product of these two improvements.

% We further validate SPARE in native vLLM serving on Granite 4.0 H-Tiny-Base using an A100 80GB GPU, where \texttt{SPARE-U32} achieves $1.41\times$ matched-batch and $2.21\times$ peak decoding throughput.
% Under online serving at concurrency 512, SPARE improves output throughput by $1.52\times$, demonstrating practical benefits beyond standalone decoding benchmarks.

\subsection{Rank-dependent Trade-offs}
\label{sec:rank_ablation}

Figure~\ref{fig:rank_ablation} examines how retained rank affects associative retrieval and decoding throughput using the original input-derived projection configurations.
Increasing rank generally improves associative memory preservation while reducing decoding throughput, revealing a clear quality--efficiency trade-off.
At matched rank budgets, uniform allocation achieves higher GPU throughput, whereas heterogeneous allocation exhibits task-dependent quality behavior.
These findings suggest that optimal recurrent-state compression must balance representation capacity, memory consumption, and GPU execution efficiency.

\subsection{Long-context Performance}
\label{sec:long_context}

Figure~\ref{fig:ruler_long_context}(a) presents six-task average RULER performance from 4K to 64K, showing that both SPARE variants retain substantial long-context capabilities despite measurable degradation relative to Dense.
At $4\times$ compression, \texttt{SPARE-H} consistently outperforms \texttt{SPARE-U}, indicating the benefits of heterogeneous state allocation.
Figure~\ref{fig:ruler_long_context}(b) further examines whether increasing the retained rank can recover long-context quality.
On 32K multi-key retrieval, increasing the mean rank from 24 to 48 improves accuracy from 85.42\% to 94.79\%, approaching the Dense baseline of 97.92\%.
At 64K, H48 also improves accuracy from 87.50\% to 90.63\%.
These results demonstrate that increasing state capacity can recover substantial long-context retrieval performance while retaining meaningful recurrent-state compression.

\subsection{Fixed-subspace Stability}
\label{sec:subspace_stability}

\begin{figure*}[t]
    \centering
    \includegraphics[width=\textwidth]{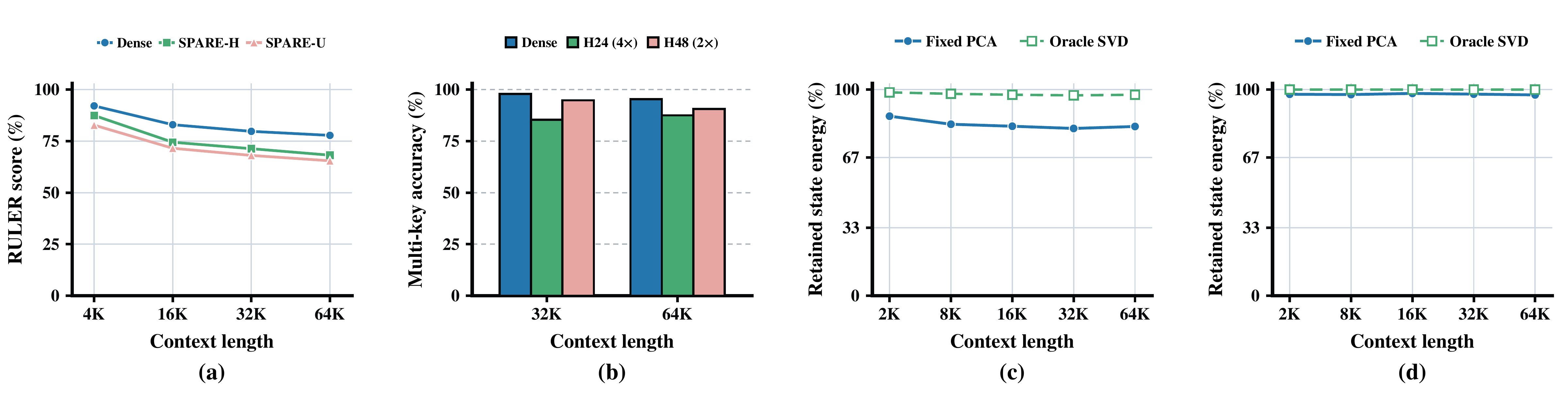}
    \caption{
    \textbf{Long-context performance and fixed-subspace stability on OLMo Hybrid 7B.}
    H24 and H48 denote heterogeneous rank allocations with average retained ranks of 24 and 48 per recurrent head, corresponding to $4\times$ and $2\times$ state compression, respectively.
    \textbf{(a)} Six-task average RULER scores from 4K to 64K, comparing Dense, SPARE-H24, and SPARE-U24 under $4\times$ compression.
    \textbf{(b)} Multi-key retrieval accuracy for Dense, SPARE-H24, and SPARE-H48 at 32K ($N=96$) and 64K ($N=64$).
    \textbf{(c)} Retained state energy of the fixed H24 subspace versus its rank-matched oracle SVD across context lengths.
    \textbf{(d)} Retained state energy of the fixed H48 subspace versus its rank-matched oracle SVD across context lengths.
    Both fixed subspaces maintain relatively stable energy retention, while H48 preserves more state energy and approaches its rank-matched oracle.
    }
    \label{fig:ruler_long_context}
\end{figure*}

We further investigate whether offline-calibrated projection bases remain effective as recurrent states evolve over long contexts.
We consider two heterogeneous rank configurations, H24 and H48, with average retained ranks of 24 and 48 per head, corresponding to $4\times$ and $2\times$ recurrent-state compression, respectively.
Figures~\ref{fig:ruler_long_context}(c--d) compare the state energy preserved by these fixed State PCA subspaces against rank-matched oracle SVD projections computed independently at each checkpoint on Dense associative-retrieval trajectories.
Across 2K--64K contexts, H24 retains approximately 81--87\% of recurrent-state energy, whereas H48 consistently preserves 97--98\%, approaching the oracle upper bound.
At 64K, increasing the average rank from 24 to 48 reduces the gap to the oracle from 15.38 to 2.57 percentage points.
Notably, energy retention remains relatively stable across context lengths, suggesting that fixed subspaces can capture persistent dominant directions of recurrent memory without online adaptation.
These findings highlight the importance of sufficient subspace capacity for preserving long-context recurrent representations.

% \subsection{Batch Capacity and Decoding Efficiency}
% \label{sec:batch_efficiency}

% Table~\ref{tab:batch_scaling} shows that $4\times$ recurrent-state compression improves both batch capacity and matched-batch decoding speed, with \texttt{SPARE-U} achieving larger gains than \texttt{SPARE-H}.
% However, increasing batch size also increases per-step decoding latency, so the resulting peak throughput improvement is substantially smaller than the product of the batch-capacity gain and matched-batch speedup.

% \begin{figure*}[t]
%     \centering
%     \includegraphics[width=\textwidth]{figures/context_length.pdf}
%     \caption{
%     \textbf{Long-context performance and fixed-subspace stability on OLMo Hybrid 7B.}
%     \textbf{(a)} Six-task average RULER scores from 4K to 64K under $4\times$ recurrent-state compression.
%     \textbf{(b)} Multi-key retrieval accuracy comparing Dense, SPARE-H24 ($4\times$), and SPARE-H48 ($2\times$) at 32K ($N=96$) and 64K ($N=64$).
%     \textbf{(c--d)} Fixed-subspace energy retention and rank-matched oracle SVD retention on associative-retrieval trajectories for H24 and H48, respectively.
%     The fixed subspaces retain a relatively stable fraction of recurrent-state energy across context lengths, while increasing rank substantially improves coverage and narrows the gap to the oracle.
%     }
%     \label{fig:ruler_long_context}
% \end{figure*}

\section{Related Work}
\label{sec:related_work}

\subsection{Linear Attention and Efficient Inference}

Linear attention replaces the growing key-value cache of conventional attention with fixed-size recurrent states, enabling memory-efficient autoregressive inference \citep{katharopoulos2020transformers}.
Subsequent architectures enhance recurrent memory through gating and delta-rule updates, including Gated Linear Attention \citep{yang2023gated}, Gated DeltaNet \citep{yang2025gated}, Kimi Delta Attention \citep{team2025kimi}, and Gated DeltaNet-2 \citep{hatamizadeh2026gated}.
Hardware-efficient implementations, such as FlashLinearAttention \citep{yang2023gated} and Lightning Attention-2 \citep{qin2024lightning}, further accelerate recurrent computation through tiled and parallel execution.
Other approaches explore specialized hardware for maintaining recurrent states on-chip and reducing memory transfers \citep{gupta2026persistent}.
Despite these advances, repeatedly accessing large recurrent states remains costly during GPU decoding, particularly under high inference concurrency.
SPARE addresses this complementary bottleneck by reducing recurrent-state storage and memory traffic through compact recurrent computation.

\subsection{Recurrent-State Compression}

Recent studies have revealed substantial low-rank redundancy in recurrent memory, with effective ranks varying across layers and attention heads \citep{nazari2026key,sun2026state,kwon2026sketchssm,pmlr-v306-menezes26a}.
To exploit this redundancy, \citet{nazari2026key} propose structured key/query channel pruning, including rank-revealing QR-based state reduction.
Similarly, \citet{sun2026state} investigate heterogeneous state-rank dynamics and introduce rank- and norm-aware pruning.
Another line of work reduces recurrent-state storage through quantization, such as LeapQuant \citep{pan2026leapquant}, which combines window-based quantization with outlier compensation.
Unlike coordinate pruning and quantization, SPARE constructs fixed low-dimensional subspaces directly from recurrent-state statistics, preserving informative representation directions beyond the original coordinates.
Furthermore, SPARE combines flexible rank allocation with native compact recurrent updates, avoiding online decomposition and dense-state reconstruction.
This formulation enables training-free recurrent-state compression across Gated DeltaNet and Mamba-2 hybrid language models while improving GPU decoding efficiency.

\section{Conclusion}
\label{sec:conclusion}

We presented \texttt{SPARE}, a training-free framework for efficient recurrent-state compression in hybrid language models.
By combining state-aware subspace construction, flexible rank allocation, and direct recurrent execution in compact representations, \texttt{SPARE} reduces memory storage and traffic without online decomposition or dense-state reconstruction.
Across Gated DeltaNet and Mamba-2 hybrid models, \texttt{SPARE} achieves $4\times$ recurrent-state compression and up to $2.00\times$ peak full-model decoding throughput, while its quality-oriented configuration preserves or improves average retrieval accuracy.
Our experiments further reveal a trade-off between heterogeneous allocation for quality preservation and uniform allocation for GPU execution efficiency, alongside task-dependent degradation on demanding long-context benchmarks.
These findings demonstrate the potential of compact recurrent memory for improving the efficiency of hybrid language model inference under practical resource constraints.

% ACL-required Limitations is conventionally unnumbered.
%\input{sections/08_limitations}

\newpage
% Acknowledgments must be omitted from anonymous review submissions.
% Uncomment only when preparing a version in which acknowledgments are allowed.
% \input{sections/09_acknowledgments}

% ACL's bibliography style is set by acl.sty.
\bibliography{references}

@inproceedings{katharopoulos2020transformers,
  title={Transformers are rnns: Fast autoregressive transformers with linear attention},
  author={Katharopoulos, Angelos and Vyas, Apoorv and Pappas, Nikolaos and Fleuret, Fran{\c{c}}ois},
  booktitle={International conference on machine learning},
  pages={5156--5165},
  year={2020},
  organization={PMLR}
}

@article{merrill2026olmo,
  title={Olmo hybrid: From theory to practice and back},
  author={Merrill, William and Li, Yanhong and Romero, Tyler and Svete, Anej and Costello, Caia and Dasigi, Pradeep and Groeneveld, Dirk and Heineman, David and Kuehl, Bailey and Lambert, Nathan and others},
  journal={arXiv preprint arXiv:2604.03444},
  year={2026}
}

@article{lieber2024jamba,
  title={Jamba: A hybrid transformer-mamba language model},
  author={Lieber, Opher and Lenz, Barak and Bata, Hofit and Cohen, Gal and Osin, Jhonathan and Dalmedigos, Itay and Safahi, Erez and Meirom, Shaked and Belinkov, Yonatan and Shalev-Shwartz, Shai and others},
  journal={arXiv preprint arXiv:2403.19887},
  year={2024}
}

@article{arora2024simple,
  title={Simple linear attention language models balance the recall-throughput tradeoff},
  author={Arora, Simran and Eyuboglu, Sabri and Zhang, Michael and Timalsina, Aman and Alberti, Silas and Zinsley, Dylan and Zou, James and Rudra, Atri and R{\'e}, Christopher},
  journal={arXiv preprint arXiv:2402.18668},
  year={2024}
}

@article{eckart1936approximation,
  title   = {The Approximation of One Matrix by Another of Lower Rank},
  author  = {Eckart, Carl and Young, Gale},
  journal = {Psychometrika},
  volume  = {1},
  number  = {3},
  pages   = {211--218},
  year    = {1936},
  doi     = {10.1007/BF02288367}
}

@inproceedings{dua-etal-2019-drop,
    title = "{DROP}: A Reading Comprehension Benchmark Requiring Discrete Reasoning Over Paragraphs",
    author = "Dua, Dheeru  and
      Wang, Yizhong  and
      Dasigi, Pradeep  and
      Stanovsky, Gabriel  and
      Singh, Sameer  and
      Gardner, Matt",
    editor = "Burstein, Jill  and
      Doran, Christy  and
      Solorio, Thamar",
    booktitle = "Proceedings of the 2019 Conference of the {N}orth {A}merican Chapter of the {A}ssociation for {C}omputational {L}inguistics: Human Language Technologies, Volume 1 (Long and Short Papers)",
    month = jun,
    year = "2019",
    address = "Minneapolis, Minnesota",
    publisher = "Association for Computational Linguistics",
    url = "https://aclanthology.org/N19-1246/",
    doi = "10.18653/v1/N19-1246",
    pages = "2368--2378"
}

@article{kwiatkowski-etal-2019-natural,
    title = "Natural Questions: A Benchmark for Question Answering Research",
    author = "Kwiatkowski, Tom  and
      Palomaki, Jennimaria  and
      Redfield, Olivia  and
      Collins, Michael  and
      Parikh, Ankur  and
      Alberti, Chris  and
      Epstein, Danielle  and
      Polosukhin, Illia  and
      Devlin, Jacob  and
      Lee, Kenton  and
      Toutanova, Kristina  and
      Jones, Llion  and
      Kelcey, Matthew  and
      Chang, Ming-Wei  and
      Dai, Andrew M.  and
      Uszkoreit, Jakob  and
      Le, Quoc  and
      Petrov, Slav",
    editor = "Lee, Lillian  and
      Johnson, Mark  and
      Roark, Brian  and
      Nenkova, Ani",
    journal = "Transactions of the Association for Computational Linguistics",
    volume = "7",
    year = "2019",
    address = "Cambridge, MA",
    publisher = "MIT Press",
    url = "https://aclanthology.org/Q19-1026/",
    doi = "10.1162/tacl_a_00276",
    pages = "452--466"
}

@inproceedings{joshi-etal-2017-triviaqa,
    title = "{T}rivia{QA}: A Large Scale Distantly Supervised Challenge Dataset for Reading Comprehension",
    author = "Joshi, Mandar  and
      Choi, Eunsol  and
      Weld, Daniel  and
      Zettlemoyer, Luke",
    editor = "Barzilay, Regina  and
      Kan, Min-Yen",
    booktitle = "Proceedings of the 55th Annual Meeting of the {A}ssociation for {C}omputational {L}inguistics (Volume 1: Long Papers)",
    month = jul,
    year = "2017",
    address = "Vancouver, Canada",
    publisher = "Association for Computational Linguistics",
    url = "https://aclanthology.org/P17-1147/",
    doi = "10.18653/v1/P17-1147",
    pages = "1601--1611"
}

@inproceedings{rajpurkar-etal-2016-squad,
    title = "{SQ}u{AD}: 100,000+ Questions for Machine Comprehension of Text",
    author = "Rajpurkar, Pranav  and
      Zhang, Jian  and
      Lopyrev, Konstantin  and
      Liang, Percy",
    editor = "Su, Jian  and
      Duh, Kevin  and
      Carreras, Xavier",
    booktitle = "Proceedings of the 2016 Conference on Empirical Methods in Natural Language Processing",
    month = nov,
    year = "2016",
    address = "Austin, Texas",
    publisher = "Association for Computational Linguistics",
    url = "https://aclanthology.org/D16-1264/",
    doi = "10.18653/v1/D16-1264",
    pages = "2383--2392"
}

@inproceedings{hao2011one,
  title={From one tree to a forest: a unified solution for structured web data extraction},
  author={Hao, Qiang and Cai, Rui and Pang, Yanwei and Zhang, Lei},
  booktitle={Proceedings of the 34th international ACM SIGIR conference on Research and development in Information Retrieval},
  pages={775--784},
  year={2011}
}

@inproceedings{ren2025samba,
  title={Samba: Simple hybrid state space models for efficient unlimited context language modeling},
  author={Ren, Liliang and Liu, Yang and Lu, Yadong and Liang, Chen and Chen, Weizhu and others},
  booktitle={International Conference on Learning Representations},
  volume={2025},
  pages={53551--53575},
  year={2025}
}

@InProceedings{pmlr-v306-menezes26a,
  title = 	 {{GHOST}: Unmasking Phantom States in Mamba2 via Grouped Hidden-state Output-aware Selection \& Truncation},
  author =       {Menezes, Michael and Kyrillidis, Anastasios},
  booktitle = 	 {Proceedings of the 43rd International Conference on Machine Learning},
  pages = 	 {87805--87821},
  year = 	 {2026},
  editor = 	 {Zhang, Tong and Dudik, Miroslav and Jaggi, Martin and Agarwal, Alekh and Li, Sharon and Schuurmans, Dale and Zhu, Jerry and Berkenkamp, Felix and Dong, Hanze and Bietti, Alberto},
  volume = 	 {306},
  series = 	 {Proceedings of Machine Learning Research},
  month = 	 {06--11 Jul},
  publisher =    {PMLR},
  url = 	 {https://proceedings.mlr.press/v306/menezes26a.html}
}

@article{kwon2026sketchssm,
  title={SketchSSM: Write to the Full State, Read from a Compact Sketch},
  author={Kwon, Omin and Shin, JoongWon and Kim, Minseo and Keutzer, Kurt and Kim, Sehoon and Lee, Jae W},
  journal={arXiv preprint arXiv:2609.33051},
  year={2026}
}

@inproceedings{NEURIPS2022_67d57c32,
 author = {Dao, Tri and Fu, Dan and Ermon, Stefano and Rudra, Atri and R\'{e}, Christopher},
 booktitle = {Advances in Neural Information Processing Systems},
 doi = {10.52202/068431-1189},
 editor = {S. Koyejo and S. Mohamed and A. Agarwal and D. Belgrave and K. Cho and A. Oh},
 pages = {16344--16359},
 publisher = {Curran Associates, Inc.},
 title = {FlashAttention: Fast and Memory-Efficient Exact Attention with IO-Awareness},
 url = {https://proceedings.neurips.cc/paper_files/paper/2022/file/67d57c32e20fd0a7a302cb81d36e40d5-Paper-Conference.pdf},
 volume = {35},
 year = {2022}
}

@article{hsieh2024ruler,
  title={RULER: What's the real context size of your long-context language models?},
  author={Hsieh, Cheng-Ping and Sun, Simeng and Kriman, Samuel and Acharya, Shantanu and Rekesh, Dima and Jia, Fei and Zhang, Yang and Ginsburg, Boris},
  journal={arXiv preprint arXiv:2404.06654},
  year={2024}
}

@article{pan2026leapquant,
  title={LeapQuant: Efficient Linear Attention with Accurate Recurrent State Quantization},
  author={Pan, Yi and Xi, Haocheng and Zhu, Kan and Li, Xingyang and Wu, Yibo and Mishra, Mayank and Zhang, Hongtao and Zheng, William X and Kasikci, Baris and Han, Song and others},
  journal={arXiv preprint arXiv:2609.38166},
  year={2026}
}

@article{hatamizadeh2026gated,
  title={Gated DeltaNet-2: Decoupling erase and write in linear attention},
  author={Hatamizadeh, Ali and Choi, Yejin and Kautz, Jan},
  journal={arXiv preprint arXiv:2605.22791},
  year={2026}
}

@article{team2025kimi,
  title={Kimi linear: An expressive, efficient attention architecture},
  author={Team, Kimi and Zhang, Yu and Lin, Zongyu and Yao, Xingcheng and Hu, Jiaxi and Meng, Fanqing and Liu, Chengyin and Men, Xin and Yang, Songlin and Li, Zhiyuan and others},
  journal={arXiv preprint arXiv:2510.26692},
  year={2025}
}

@article{qin2024lightning,
  title={Lightning attention-2: A free lunch for handling unlimited sequence lengths in large language models},
  author={Qin, Zhen and Sun, Weigao and Li, Dong and Shen, Xuyang and Sun, Weixuan and Zhong, Yiran},
  journal={arXiv preprint arXiv:2401.04658},
  year={2024}
}

@inproceedings{yang2025gated,
  title={Gated delta networks: Improving mamba2 with delta rule},
  author={Yang, Songlin and Kautz, Jan and Hatamizadeh, Ali},
  booktitle={International Conference on Learning Representations},
  volume={2025},
  pages={29687--29707},
  year={2025}
}

@article{sun2026state,
  title={State Rank Dynamics in Linear Attention LLMs},
  author={Sun, Ao and Zhang, Hongtao and Zhou, Heng and Ma, Yixuan and Qin, Yiran and Su, Tongrui and Liu, Yan and Ma, Zhanyu and Xu, Jun and Gao, Jiuchong and others},
  journal={arXiv preprint arXiv:2602.02195},
  year={2026}
}

@article{nazari2026key,
  title={The Key to State Reduction in Linear Attention: A Rank-based Perspective},
  author={Nazari, Philipp and Rusch, T Konstantin},
  journal={arXiv preprint arXiv:2602.04852},
  year={2026}
}

@article{dao2024transformers,
  title={Transformers are SSMs: Generalized models and efficient algorithms through structured state space duality},
  author={Dao, Tri and Gu, Albert},
  journal={arXiv preprint arXiv:2405.21060},
  year={2024}
}

@article{gupta2026persistent,
  title={A Persistent-State Dataflow Accelerator for Memory-Bound Linear Attention Decode on FPGA},
  author={Gupta, Neelesh and Wang, Peter and Kannan, Rajgopal and Prasanna, Viktor K},
  journal={arXiv preprint arXiv:2603.05931},
  year={2026}
}

@article{vaswani2017attention,
  title={Attention is all you need},
  author={Vaswani, Ashish and Shazeer, Noam and Parmar, Niki and Uszkoreit, Jakob and Jones, Llion and Gomez, Aidan N and Kaiser, {\L}ukasz and Polosukhin, Illia},
  journal={Advances in neural information processing systems},
  volume={30},
  year={2017}
}

@article{yang2023gated,
  title={Gated linear attention transformers with hardware-efficient training},
  author={Yang, Songlin and Wang, Bailin and Shen, Yikang and Panda, Rameswar and Kim, Yoon},
  journal={arXiv preprint arXiv:2312.06635},
  year={2023}
}

% Supplementary appendices use lettered headings (A, B, ...).

\newpage
\appendix

\end{document}